\documentclass[sigconf,nonacm]{acmart}
\AtBeginDocument{%
  }

\copyrightyear{2026}
\acmYear{2026}

\usepackage{tcolorbox}
\usepackage{listings}
\usepackage{booktabs}
\usepackage{multirow}
\begin{document}

\title{AdaHome: An Adaptive Smart Home Assistant using Local Small Language Models}
\titlenote{This paper is an extended version of: Eu Jin Lim, Zhaoxing Li, and Sebastian Stein. 2026. AdaHome: An Adaptive Smart Home Assistant using Local Small Language Models. In \textit{Proceedings of the International Conference on Multimodal Interaction (ICMI '26), October 05--09, 2026, Napoli, Italy}. ACM, New York, NY, USA. \url{https://doi.org/10.1145/3776574.3831158}}

\author{Eu Jin Lim}
\affiliation{%
  \institution{University of Southampton}
  \city{Southampton}
  \country{United Kingdom}}
\email{ejl1e22@soton.ac.uk}

\author{Zhaoxing Li}
\affiliation{%
  \institution{University of Southampton}
  \city{Southampton}
  \country{United Kingdom}}
\email{zhaoxing.li@soton.ac.uk}

\author{Sebastian Stein}
\affiliation{%
  \institution{University of Southampton}
  \city{Southampton}
  \country{United Kingdom}}
\email{ss2@ecs.soton.ac.uk}


\begin{abstract}
Smart home assistants interpret a wide range of user commands, from explicit device control to underspecified and preference dependent requests.  While recent systems based on Large Language Models (LLMs) improve this capability, they often rely on heavyweight reasoning pipelines and cloud-based deployment, limiting their efficiency and suitability for resource-constrained environments, and raising privacy concerns. In addition, existing approaches provide limited support for stable long-term personalization. To address these issues, we present AdaHome, an adaptive smart home assistant designed for locally deployed small language models in smart home environments. Rather than applying complex reasoning uniformly, AdaHome introduces an intent-aware planning framework that dynamically routes commands either to straightforward prompt-based or lightweight reasoning-based components. For commands requiring interpretation, we adopt a Chain-of-Draft strategy to enable efficient and stable decision-making. To support personalization, we further propose a preference adaptation mechanism that learns from user feedback over time without requiring prompt augmentation or model retraining. We evaluate AdaHome against representative LLM-based baselines under a unified small model setting. AdaHome achieves substantially higher accuracy on direct commands (86.7\%) while reducing latency by up to 3$\times$. Furthermore, it maintains competitive performance on ambiguous inputs with lower computational cost. In multi-turn scenarios, AdaHome achieves 88\% preference consistency, compared to 52.5\% for a prompt augmentation baseline.

\end{abstract}




\maketitle

\section{Introduction}

Interacting with smart-home environments using natural language remains challenging. Users often express commands in terms of goals, contexts, or preferences. For instance, ``make the room cozy'' or ``I want to relax'' \cite{king2024sasha}. However, smart home systems require explicit instructions that specify which devices to control and what actions to perform. This creates a fundamental gap between how users communicate and how systems execute commands. Bridging this gap requires inferring actions from inputs that are underspecified and ambiguous. The integration of Large Language Models (LLMs) has improved this capability \cite{gallo2024conversational, king2023get}. LLMs can reason over contextual information and generate structured action plans, supporting more flexible interaction beyond traditional rule-based pipelines \cite{king2024sasha, shi2024bridging, rivkin2024aiot}.

Despite these advances, significant challenges remain for practical deployment. Existing LLM-based smart home systems rely on large models and computationally intensive reasoning pipelines. These are typically deployed in cloud-based settings. While effective, such designs introduce latency, require substantial computational resources, and raise privacy concerns \cite{yin2025harmony, huang2026towards}. In real-world smart homes, responsiveness and data privacy are major concerns, making locally deployed models highly desirable. However, operating under local resource constraints requires the use of Small Language Models (SLMs). The reduced reasoning capability of SLMs presents additional challenges in maintaining reliability and performance. Furthermore, current approaches provide limited support for long-term personalization. Although LLMs can incorporate short-term conversational context, they generally lack mechanisms to continually adapt to user preferences across interactions. As a result, ambiguous commands are interpreted in a generic manner, failing to capture individual habits or evolving user behaviour. Supporting continual preference adaptation without costly model retraining remains an open challenge.

Motivated by these limitations, we present \textbf{AdaHome}, an adaptive smart home assistant designed for local deployment with small language models. AdaHome is based on the observation that not all commands require the same level of reasoning. Direct commands can be executed with minimal processing, while indirect or ambiguous inputs benefit from additional reasoning. To this end, AdaHome introduces an intent-aware planning framework that selectively applies lightweight reasoning based on command category. This allows the system to improve efficiency while maintaining action accuracy. Beyond that, AdaHome incorporates a novel preference adaptation mechanism for continual personalization. Instead of relying on prompt augmentation or model retraining, AdaHome builds a lightweight preference memory from confirmed and corrected interactions. It uses this memory to estimate future device choices, weighting past interactions based on semantic similarity and temporal decay. This allows the system to adapt incrementally to user behaviour while remaining robust to temporary deviations.

In summary, our contributions are as follows:
\begin{itemize}
    \item We propose \textbf{AdaHome}, an adaptive smart home assistant that enables efficient and personalized control using small language models.
    
    \item We introduce an \textbf{intent-aware planning framework} that selectively applies lightweight reasoning based on command category, improving efficiency while maintaining action accuracy.
        
    \item We develop a \textbf{continual preference adaptation mechanism} that enables personalized behaviour from user interactions without requiring prompt augmentation or model retraining.
    
\end{itemize}




\section{Related Work}

\subsection{LLM-based Smart Home Assistants}
Recent work has explored the use of Large Language Models (LLMs) to improve natural language interaction in smart home environments. Early studies demonstrate how LLMs can assist smart spaces by interpreting high-level user intent and generating appropriate device actions~\cite{king2023get, gallo2024conversational, shi2024bridging}. Building on this direction, more recent work models smart home control as a reasoning problem. The Sasha system~\cite{king2024sasha} formulates this as a goal-oriented reasoning problem, decomposing user intent into intermediate steps before execution. This structured design improves action relevance and reduces false positives. However, it does not support personalization. Subsequent approaches extend this towards agent-based designs. For instance, SAGE \cite{rivkin2024aiot} explores the use of autonomous LLM-based agents for smart-home control, enabling more flexible handling of complex tasks. However, this flexibility comes at the cost of heavier reasoning pipelines and increased computational overhead. Despite these advances, most existing systems depend on large models like GPT-4 and computationally intensive reasoning processes, which are typically deployed in cloud-based settings. This limits their practicality for local, resource-constrained deployments. Furthermore, reliance on cloud services introduces privacy risks, such as potential data breaches or unauthorised access during data transmission~\cite{wang2025comprehensive}.

This has motivated recent efforts towards the use of Small Language Models (SLMs) in smart home environments. Harmony~\cite{yin2025harmony} introduces a modular architecture powered by a locally deployed SLM. By enforcing structured reasoning and incorporating a rule-based controller for action validation, the system improves reliability and mitigates hallucinations under limited model capacity. Similarly, HomeLLaMA~\cite{huang2026towards} adopts a hybrid approach that combines offline SLM adaptation with online interaction. These approaches demonstrate the feasibility of building locally deployed smart home assistants using SLMs. However, existing systems typically apply complex reasoning pipelines uniformly across all commands, regardless of their complexity. This can lead to unnecessary computational overhead, especially for simple instructions that do not require multi-step reasoning. Prior work has also reported that such complex reasoning pipelines can incur substantial latency even for straightforward commands~\cite{yu2026leveraging}. This highlights the need for more efficient and adaptive mechanisms that tailor reasoning to the nature of the user request.

\subsection{Long-term Personalization}
Long-term personalization is an important feature for smart home assistants, as user preferences can vary across individuals. Although recent LLM-based systems demonstrate strong capabilities in interpreting user intent, continual adaptation to user preferences remains underexplored. Some prior work incorporates personalization through interaction history. For instance, SAGE~\cite{rivkin2024aiot} leverages past interactions to guide future decisions by retrieving relevant historical context via Retrieval-Augmented Generation (RAG) \cite{zhu2025large}. Similarly, IoTGPT~\cite{yu2026leveraging} abstracts user preferences into structured representations based on environmental properties and incorporates them into the model context through keyword-based matching. While these approaches enable a degree of personalization, they rely on prompt-level augmentation. As the interaction history grows, the amount of appended context increases, making it difficult to maintain relevance and efficiency within the limited context capacity of small language models. More fundamentally, such approaches struggle to balance the stability–plasticity trade-off, where the system must retain previously learned preferences (stability) while adapting to new user behaviour (plasticity) \cite{lai2025pareto, zhang2025survey}.

Beyond that, preference learning has also been explored through model-level adaptation, where models are fine-tuned to align with user preferences \cite{lismart}. These approaches are computationally intensive and not suitable for resource-constrained environments, as updates to user preferences typically require model retraining. Overall, this highlights the lack of efficient mechanisms for continual preference adaptation without reliance on prompt-level augmentation or model retraining.

\section{System Design}
\label{sec:system_design}

\begin{figure*}[h]
  \centering
  \includegraphics[width=0.7\textwidth]{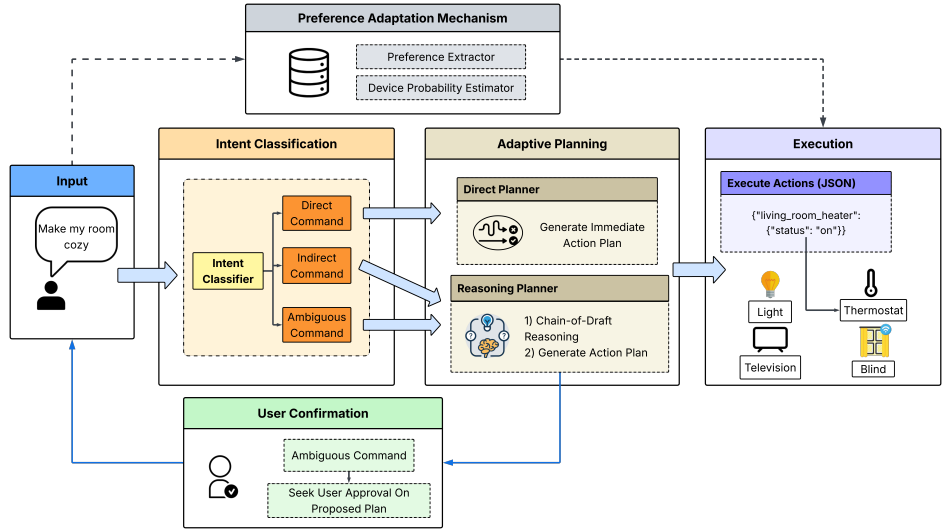}
  \caption{Overview of AdaHome system architecture, consisting of an intent-aware planning pipeline and a continual preference adaptation module. The pipeline routes user commands to either a direct or reasoning planner based on command category, while the preference module leverages past interactions to enable personalized responses.}
  \label{fig:system_architecture}
  \Description{xxx}
\end{figure*}

Figure~\ref{fig:system_architecture} shows the architecture of AdaHome, which consists of an intent-aware planning pipeline and a continual preference adaptation module. The pipeline routes inputs to either a direct planner or a reasoning planner based on command category, while the preference module leverages semantically similar past interactions to enable personalized responses.

\paragraph{Intent Classification}

\begin{table}[t]
  \caption{Categorization of user commands.}
  \label{tab:intent_categories}
  \resizebox{\columnwidth}{!}{%
  \begin{tabular}{ccl}
    \toprule
    \textbf{Category} & \textbf{Description} & \textbf{Example} \\
    \midrule
    Direct & Explicit device and action & Turn on bedroom light \\
    Indirect & Implicit intent from context & The room is too hot \\
    Ambiguous & Preference-dependent intent & I want to relax \\
    \bottomrule
  \end{tabular}%
  }
\end{table}

We first classify each user command into one of three categories based on its level of specification. As shown in Table~\ref{tab:intent_categories}, commands are categorized as \textit{direct}, \textit{indirect}, or \textit{ambiguous}, and the assigned category determines the subsequent planning strategy. Direct commands explicitly specify both the target device and action. Indirect commands express intent implicitly but can be resolved using the current device context. In contrast, ambiguous commands are preference-dependent and may admit multiple valid interpretations across users. The classifier is implemented as a structured LLM prompt (see Appendix~\ref{app:intent_classification}) that outputs a category label. Its impact on downstream performance is evaluated in Section~\ref{sec:results}.

\paragraph{Adaptive Planning}

Following intent classification, AdaHome routes each command to an appropriate planner, applying reasoning only when necessary. Direct commands are handled by a direct planner using a straightforward prompt without intermediate reasoning, enabling low-latency execution. Indirect and ambiguous commands are processed by a reasoning planner, which interprets user intent in the context of available devices to generate an action plan. To ensure reliability, both planners are constrained by a predefined device schema (Appendix~\ref{app:smart_home_setting}). Generated actions are validated against valid device names and supported attributes defined in the schema. Invalid outputs are corrected through constrained reprompting while preserving the intended action. This schema-guided process reduces hallucinations and ensures consistent, executable outputs.

\paragraph{User Confirmation}

For ambiguous commands, AdaHome introduces a user confirmation step before execution. The generated action plan is presented for approval or correction, reducing errors from multiple valid interpretations and providing an additional safeguard for sensitive operations such as door unlocking or blind control. This improves both safety and reliability. User feedback also serves as supervision for continual preference adaptation. Confirmed or corrected actions are recorded in preference memory (detailed in Section~\ref{sec:user_pref}), enabling the system to better align with user-specific behaviour over time and reducing the need for repeated confirmations.

\paragraph{Execution}

Once validated and confirmed, the final action plan is represented as structured JSON and dispatched to the smart home environment through an execution controller that maps commands to the corresponding device APIs.

To support efficient decision-making and personalization under resource constraints, AdaHome combines a lightweight reasoning strategy with a dedicated mechanism for continual preference adaptation. In the following, we describe the reasoning approach (Section~\ref{sec:cod}) and the preference adaptation mechanism (Section~\ref{sec:user_pref}).

\subsection{Chain-of-Draft Reasoning}
\label{sec:cod}
To handle commands requiring interpretation, AdaHome adopts a lightweight reasoning strategy within the reasoning planner. Common reasoning approaches such as Chain-of-Thought (CoT)~\cite{wei2022chain} and ReAct-style prompting~\cite{yao2022react} generate detailed intermediate reasoning or involve multiple interaction steps, leading to increased token usage and latency. This makes them less suitable for resource-constrained smart home systems. In addition, small language models (SLMs) have limited reasoning capacity and may overgenerate unnecessary interpretations even for simple commands, causing unstable outputs~\cite{han2025enhancing, li2025small}. 

To address this, we employ Chain-of-Draft (CoD) prompting~\cite{xu2025chain}, which encourages the model to produce a minimal intermediate representation sufficient for decision making (see Appendix~\ref{app:cod}). Given an input, the model generates a compact reasoning draft capturing the inferred intent, followed by a structured JSON action specifying device state changes. Both the draft and action are generated in a single pass. This design reflects the observation that smart home control primarily requires mapping user intent to executable actions rather than generating detailed explanations. As a result, verbose reasoning introduces additional latency without improving decision quality. By constraining reasoning to a short draft, CoD reduces token usage while preserving the essential semantic information required for accurate planning.

\begin{figure}[h]
  \centering
  \includegraphics[width=\linewidth]{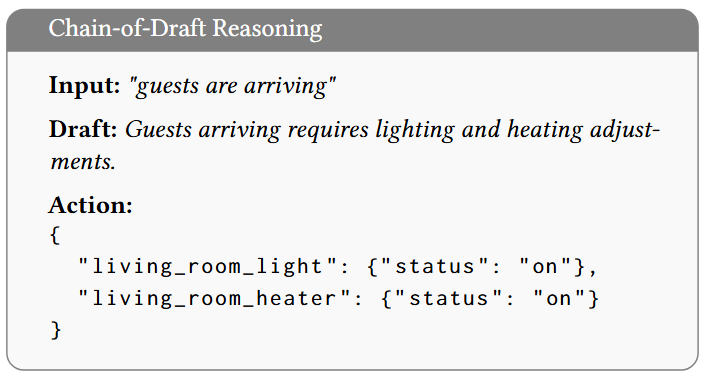}
  \caption{An example of Chain-of-Draft reasoning generating a JSON action.}
  \label{fig:cod_example}
  \Description{A text box showing a Chain-of-Draft example. Input is guests are arriving. Draft explains lighting and heating adjustments are needed. Action shows a JSON code block turning on the living room light and heater.}
\end{figure}

Furthermore, SLMs are particularly sensitive to long and verbose reasoning outputs, which can lead to inconsistency and hallucination~\cite{han2025enhancing, li2025small}. The compact structure of CoD improves output stability and reliability, making it well-suited for on-device deployment. Figure~\ref{fig:cod_example} illustrates an example of CoD reasoning and its output. The draft provides a concise abstraction of the inferred intent, while the JSON output directly specifies executable device actions.

\subsection{User Preference Adaptation}
\label{sec:user_pref}

User preference modeling is challenging, because it is highly contextual and dynamic. For example, a user issuing the command ``make the room comfortable'' may typically prefer bright lighting, but under certain circumstances may instead prefer dim lighting. Such deviations are often temporary, and the user may revert to their usual preference in subsequent interactions. A system that relies primarily on preference retrieval (e.g., \cite{rivkin2024aiot, yu2026leveraging}) would struggle to capture this behaviour~\cite{zhang2025survey}. This motivates modeling preferences based on accumulated interaction history.

AdaHome incorporates a lightweight preference adaptation mechanism that operates without model retraining or prompt augmentation, making it suitable for locally deployed small language models. Each confirmed or corrected action is treated as a preference signal. After user confirmation, a preference extractor extracts the underlying intent from the user command in one or two phrases (see Appendix~\ref{app:preference_extraction}), filtering out irrelevant context. This normalized representation ensures that semantically equivalent commands (e.g., ``movie night'' and ``let's watch a movie'') map to similar embeddings, improving the reliability of similarity-based retrieval. The extracted intent is then encoded into an embedding vector using a sentence embedding model. This vector is indexed in a local vector database, while the corresponding timestamp and resulting device states are stored in a local relational database. In this work, preference adaptation is formulated over binary device states ($x_i \in \{0,1\}$), where $1$ denotes activation and $0$ denotes deactivation. Continuous parameters (e.g., temperature) are not considered within this formulation, as modeling such attributes requires a different representation.

\begin{figure*}[h]
  \centering
  \includegraphics[width=0.7\textwidth]{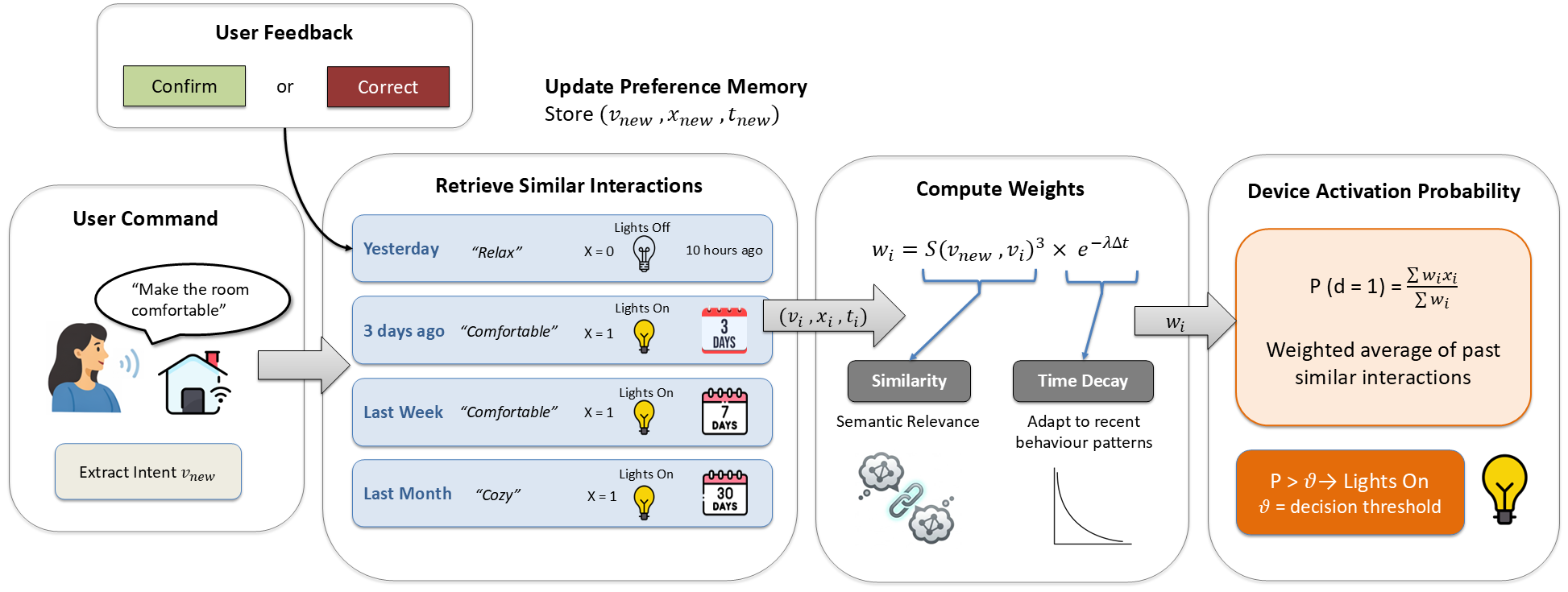}
  \caption{
    Overview of AdaHome’s preference adaptation mechanism. The system retrieves semantically similar past interactions, computes weights, and estimates device activation probabilities. User feedback, when available, is incorporated to update the preference memory. 
    }
  \label{fig:user_pref}
  \Description{}
\end{figure*}

As illustrated in Figure~\ref{fig:user_pref}, during inference the system retrieves the top-$k$ most semantically similar past interactions (we use $k=10$) using FAISS nearest neighbor search \cite{douze2024faiss} in the embedding space. A cosine similarity threshold is then applied to filter out weak matches, ensuring that only relevant interactions are retained for prediction. If no interactions satisfy this similarity threshold, the system proceeds with the intent-aware planning pipeline as described in Section~\ref{sec:system_design}. Given a new command, the extracted intent is embedded as $v_{\text{new}}$. Each retrieved memory $i$ consists of a historical embedding $v_i$, a binary device state $x_i$, and a timestamp $t_i$. 

To estimate how strongly each previous interaction should influence the current decision, we compute a semantic time decay weight $w_i$: 

\begin{equation}
w_i = S(v_{\text{new}}, v_i)^\gamma \cdot e^{-\lambda \Delta t_i}
\end{equation}

where $S(\cdot,\cdot)$ denotes cosine similarity, $\Delta t_i$ (in days) is the elapsed time since interaction $i$ was recorded, and $\lambda$ controls the rate at which older interactions lose influence. The similarity term captures semantic relevance between the current and past intents, while $\gamma$ controls the sharpness of the weighting, increasing the contrast between highly similar and weakly similar interactions. In our implementation, we fix $\gamma = 3$, which increases selectivity without overly concentrating the weighting on a single interaction. The exponential term prioritizes more recent interactions, consistent with temporal forgetting patterns in memory modeling~\cite{zhong2024memorybank}. Here we set $\lambda = 0.1$, which reduces the influence of a past interaction to approximately 0.5 after 7 days. This timescale aligns with recurring weekly patterns observed in smart home usage~\cite{salama2025development}. As a result, older or infrequent interactions continue to contribute with diminishing weights.

The probability that a device $d$ should be activated is then computed using a kernel regression estimator \cite{nadaraya1964estimating, watson1964smooth}:
\begin{equation}
P(d=1) = \frac{\sum_i w_i x_i}{\sum_i w_i}
\end{equation}

where the weights act as the kernel function over the embedding space. It estimates the likelihood of activating a device based on the weighted average of similar past interactions. Interactions that are both semantically similar and more recent contribute more strongly to the final decision. Over time, frequent behaviours accumulate and dominate the prediction, making the system robust to occasional deviations. At the same time, if the user consistently changes their behaviour, the newly accumulated preferences will gradually shift the decision boundary. 

Finally, user feedback is incorporated through confirmation or correction, which is stored as a new preference tuple $(v_{\text{new}}, x_{\text{new}}, t_{\text{new}})$ and added to the memory. This creates a continuous feedback loop that allows AdaHome to refine its behaviour over time without requiring retraining.

\section{Evaluation}

We conduct a quantitative evaluation to assess AdaHome based on two aspects. First, we examine the system's ability to translate diverse user instructions into structured device actions. Second, we assess its capability to support continual user preference adaptation over time. Our evaluation adopts a unified small model setting to reflect realistic edge deployment constraints. All systems are implemented using the same lightweight model (Llama 3.2-3B~\cite{grattafiori2024llama}). For all experiments, Ollama\footnote{\url{https://ollama.com}} is used as the inference engine, and models are executed using greedy decoding (temperature = 0.0) to minimize randomness. Experiments are conducted in a controlled environment defined by a fixed smart home device schema which specifies available devices, current states, and allowable actions. This ensures consistent validation of generated outputs across systems. To reflect resource-constrained edge deployment, all systems are executed locally on a machine equipped with an AMD Ryzen 5 220 CPU and 16GB RAM, representative of a low-cost consumer device (approximately £200-£300). 

\subsection{Comparative Study}
We compare all systems in a single-turn interaction setting, where memory access to prior interaction history and user feedback are disabled. This isolates the core decision-making behaviour of each system and removes any external factors. Each experiment is repeated three times to ensure consistency.

\subsubsection{Experimental Setup}
\label{sec:experiment}
\paragraph{Smart Home Setting}
We establish a standardized smart home environment with 12 predefined devices spanning multiple functional categories such as lighting, climate control, appliances, and security. Each device is associated with a fixed set of controllable attributes. A detailed device schema is provided in Appendix~\ref{app:smart_home_setting}.


\paragraph{Baselines}

We compare AdaHome against three representative LLM-based smart home baselines. We focus on evaluating their architectural behaviour under a unified small model setting. This setting reflects practical edge deployment scenarios and enables a controlled analysis of how different system designs operate under limited model capacity. Each baseline is reproduced to preserve its core architectural design while operating under identical model and environment constraints.

\textbf{Sasha}~\cite{king2024sasha} is a goal-oriented LLM smart home assistant that decomposes instruction processing into feasibility checking, device filtering, and action planning. We reproduce Sasha following this multi-stage pipeline detailed in the original paper, with the interactive feedback loop disabled for consistency with our automated evaluation.

\textbf{SAGE}~\cite{rivkin2024aiot} is an agent-based system that performs iterative tool-based reasoning. We reimplement its coordinator and tool-selection loop within our fixed device schema, focusing on device control tasks and excluding auxiliary capabilities (e.g., web search and visual inputs). 

\textbf{Harmony}~\cite{yin2025harmony} is a modular smart home assistant designed for locally deployed lightweight models. We reproduce its three-stage pipeline of message handling, grounded reasoning, and rule-based validation, with memory components disabled.

\paragraph{Dataset}
We construct an evaluation dataset using commands derived from the Sasha dataset (40 commands) and the SAGE dataset (50 commands), retaining only those compatible with our predefined device schema. We further extend the dataset with manually constructed commands to ensure balanced coverage across the three categories in Table~\ref{tab:intent_categories}. The final dataset contains 90 commands: 30 direct, 30 indirect, and 30 ambiguous. The full dataset is provided in Appendix~\ref{app:dataset}.

\paragraph{Evaluation Metrics}
We evaluate all systems along two dimensions: action accuracy and computational efficiency. For direct commands, where the desired outcome is explicitly specified, we use a \textit{binary success rate} based on exact matching of the final device state. To further characterize error types, we additionally report \textit{false positive (FP)} and \textit{false negative (FN)} rates, where FP indicates the presence of any extra activated device, and FN indicates failure to activate required devices.

For indirect and ambiguous commands, strict exact matching is overly restrictive because the desired outcome is not fully specified. Indirect commands typically imply a primary actionable device (e.g., ``the living room is dusty'' suggesting activation of a robot vacuum), but additional actions like ``turning on air purifier'' are also reasonable and should not be penalized. Ambiguous commands are inherently preference-dependent, so multiple valid action plans exist. For example, a ``morning routine" may correspond to different device combinations across users. In both cases, we evaluate outputs based on semantic appropriateness rather than exact matching. Specifically, we employ an \textit{LLM-as-Judge} \cite{zheng2023judging} to assess whether the generated action outcome is appropriate given the command and device state. To reduce evaluation bias, the judge uses a larger model from a different family (Qwen2.5-14B~\cite{qwen2025qwen25technicalreport}) with a fixed evaluation prompt (see Appendix~\ref{app:llm_judge_prompt}), and outputs a binary success/fail decision. Example outputs are shown in Appendix~\ref{app:llm_judge_example}.

To assess the reliability of the LLM-based evaluation, we query the judge three times and use majority voting. We further compare its outputs against human annotations on a stratified subset of 48 examples spanning all systems and command categories. The human annotations were performed independently by members of the research team, following the same evaluation criteria provided to the LLM judge. The LLM-Judge achieves substantial agreement with human evaluation (Cohen’s $\kappa = 0.834$), indicating that it provides a reliable proxy for human assessment in our setting.

For computational efficiency, we measure end-to-end inference latency and report the number of input and output tokens per interaction.

\subsubsection{Results}
\label{sec:results}
\begin{figure}[h]
    \centering
    \includegraphics[width=0.7\columnwidth]{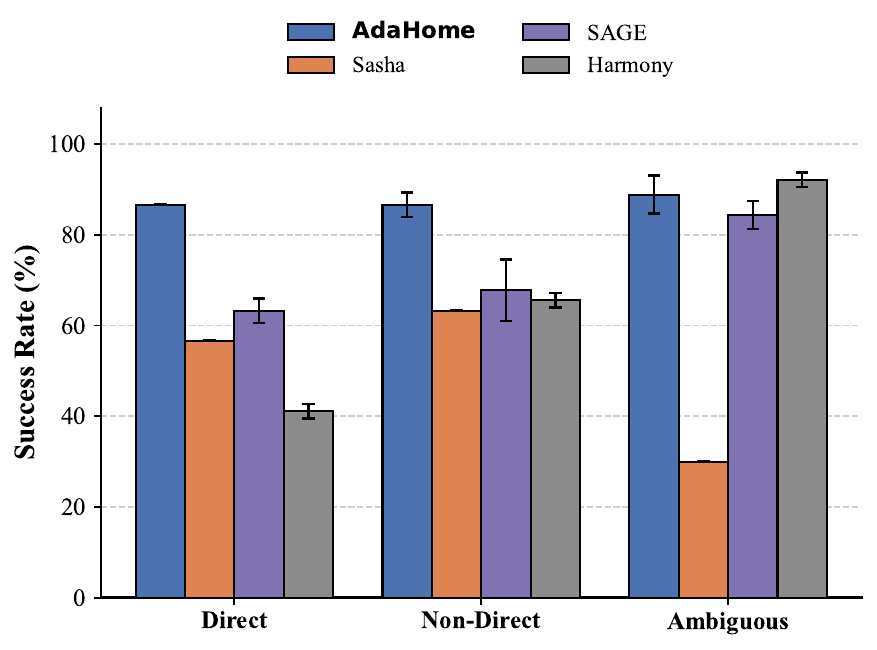}
    \caption{Success rate across Direct, Indirect, and Ambiguous command types. Error bars denote standard deviation over three runs. In cases where they are not visible, the variance is negligible.}
    \label{fig:single_turn_results}
    \Description{xx}
\end{figure}

The aggregated results are presented in Table~\ref{tab:main_results}, with a summary visualization shown in Figure~\ref{fig:single_turn_results}.

\begin{table*}[t]
\centering
\caption{Comparative study results averaged over three runs. For Indirect and Ambiguous commands, FP and FN are not applicable (-) as success is determined via goal satisfaction. Best results in each category are bolded.}
\label{tab:main_results}
\resizebox{0.8\textwidth}{!}{%
\begin{tabular}{ll ccc ccc}
\toprule
\textbf{Category} & \textbf{System} & \textbf{Success (\%)} & \textbf{FP (\%)}& \textbf{FN (\%)} & \textbf{Latency (s)} & \textbf{Input Tokens} & \textbf{Output Tokens}  \\
\midrule
\multirow{4}{*}{\textbf{Direct}} 
& Sasha & 56.7 & 0.0 & 43.3  & 15.07 & 619 & 58 \\
& SAGE & 63.3 & 2.2 & 36.7 & 17.95 & 757 & 105 \\
& Harmony & 41.1 & 37.8 & 31.1 & 28.84 & 799 & 168  \\
& \textbf{AdaHome} & \textbf{86.7} & 0.0 & 13.3 & \textbf{10.48} & 441 & 32 \\
\midrule
\multirow{4}{*}{\textbf{Indirect}} 
& Sasha & 63.3 & - & - & 17.47 & 659 & 67 \\
& SAGE & 67.8 & - & - & 17.45 & 660 & 94  \\
& Harmony & 65.6 & - & - & 30.71 & 812 & 171 \\
& \textbf{AdaHome} & \textbf{86.7} & - & - & \textbf{13.28} & 470 & 50  \\
\midrule
\multirow{4}{*}{\textbf{Ambiguous}} 
& Sasha & 30.0 & - & - & 14.87 & 463 & 74 \\
& SAGE & 84.4 & - & - & 19.16 & 660 & 114 \\
& Harmony & \textbf{92.2} & - & - & 32.93 & 819 & 195 \\
& \textbf{AdaHome} & 88.9 & - & - & \textbf{14.47} & 463 & 65  \\
\bottomrule
\end{tabular}%
}
\end{table*}

\paragraph{Action Accuracy}
AdaHome achieves the highest success rate on direct commands (86.7\%), outperforming all baselines. Harmony exhibits a high false positive rate (37.8\%) because its mandatory reasoning stage tends to over-interpret explicit instructions, leading to spurious device activations. Conversely, Sasha suffers from a high false negative rate, as relevant devices are often excluded during intermediate feasibility checking and filtering. Similarly, SAGE relies on iterative tool-based reasoning, which introduces a dependency on multi-step decision-making and termination control. Under small model constraints, the system may repeatedly invoke intermediate tools (e.g., capability retrieval) without progressing to final action execution. These results reveal a key limitation of applying complex reasoning pipelines to Small Language Models (SLMs). When the task is already well-specified, additional reasoning increases the likelihood of hallucination or intent deviation. AdaHome mitigates this by using a straightforward, instruction-focused prompt combined with a schema-guided generation process, improving execution reliability under limited model capacity.

A similar pattern is observed for indirect commands. AdaHome maintains the highest success rate (86.7\%), while Sasha, SAGE and Harmony achieve comparable performance in the range of approximately 60--70\%. In this category, AdaHome employs a lightweight Chain-of-Draft (CoD) reasoning strategy to resolve implicit intent. Unlike heavy reasoning pipelines, CoD provides a concise intermediate representation that guides action generation without introducing unnecessary inference steps. This allows the model to capture implicit intent while avoiding the over-generation and error propagation observed in more complex reasoning frameworks. The results indicate that structured, minimal reasoning is sufficient for resolving indirect commands under SLM constraints.

For ambiguous commands, a different trend emerges. Harmony achieves the highest success rate (92.2\%), followed by AdaHome (88.9\%) and SAGE (84.4\%). This reflects the advantage of more expressive reasoning pipelines in exploring a wider space of possible interpretations for underspecified inputs. In contrast, AdaHome prioritizes efficiency through its lightweight CoD strategy, which may limit exhaustive exploration but still achieves competitive performance. Notably, Sasha performs poorly (30.0\%), largely due to its conservative design that abstains when the task cannot be confidently resolved.

\paragraph{Impact of Intent Classification}
Because AdaHome's performance relies on dynamically routing commands to the correct planner, we further examine the impact of intent classification errors. The intent classifier achieves an accuracy of 75.6\% on the evaluation dataset. Direct commands are occasionally misclassified as indirect, which routes some direct inputs to the reasoning planner. While this introduces additional latency, it generally preserves correct action generation. For indirect and ambiguous commands, routing errors mainly affect whether a confirmation step is triggered, as both cases are handled by the same reasoning planner. Overall, misclassification affects execution efficiency rather than action correctness, indicating that the system is robust to routing errors.

\begin{figure}[h]
    \centering
    \includegraphics[width=0.7\columnwidth]{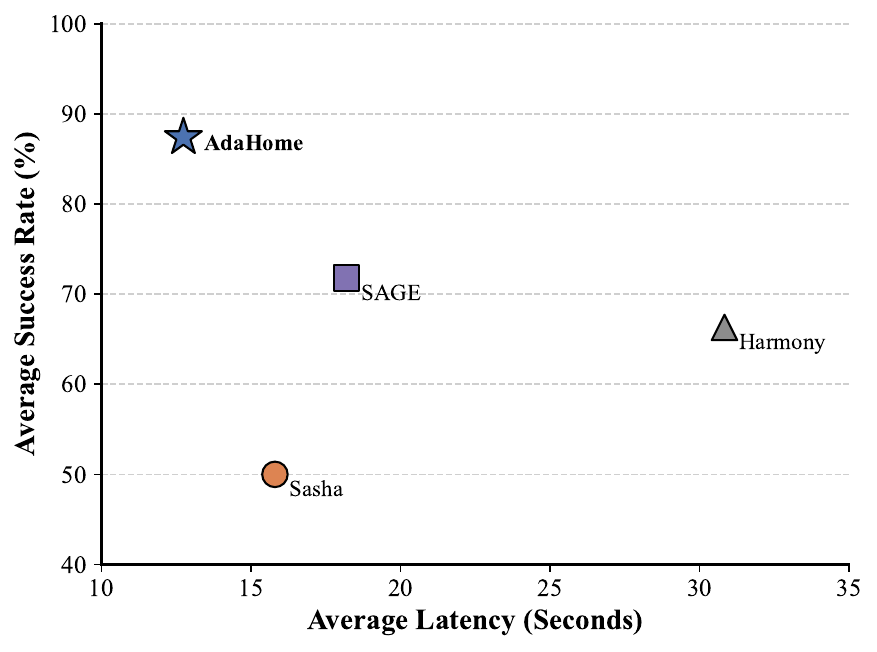}
    \caption{Overall trade-off between average end-to-end latency and average success rate across all command categories.}
    \label{fig:scatter}
    \Description{xx}
\end{figure}

\paragraph{Computational Efficiency}
AdaHome achieves the lowest end-to-end latency and smallest token footprint across all systems and command categories. This efficiency stems from its intent-aware routing strategy, which invokes reasoning only when needed. Harmony incurs the highest latency and token usage due to verbose reasoning and additional rule-based validation, while SAGE is slowed by its iterative tool-based reasoning loop. Overall, the results reveal a clear trade-off between reasoning complexity and computational efficiency. Although more expressive reasoning pipelines can improve performance in ambiguous scenarios, they incur substantial overhead and are less robust under small model constraints. AdaHome demonstrates that a lightweight and selectively applied reasoning strategy can achieve a more favorable balance between accuracy and efficiency. This trade-off is further illustrated in Figure~\ref{fig:scatter}.

\subsection{Preference Adaptation Analysis}
We evaluate AdaHome in a multi-turn setting to assess its ability to support continual user preference adaptation over repeated interactions. Unlike the single-turn evaluation, this experiment captures evolving user behaviour across interactions. The objective is to measure whether the system can maintain consistent preferences, recover from temporary deviations, and adapt to preference change.

\paragraph{Baselines}
We compare AdaHome against a Retrieval-Augmented Generation (RAG) baseline as a representative prompt augmentation approach~\cite{lewis2020retrieval, rivkin2024aiot}. In this setup, past interactions are stored and retrieved based on semantic similarity to the current user instruction. During inference, the retrieved preferences are incorporated into the model context. When user preferences change, the stored memory is updated accordingly. To ensure a fair comparison, the RAG baseline uses the same model prompt, with Chain-of-Draft reasoning applied.

\paragraph{Dataset}
We construct a longitudinal dataset using ambiguous commands from the comparative study to simulate repeated user interactions over time (see Appendix~\ref{app:pref_dataset}). The dataset consists of 30 sequences, each containing 8 interaction turns. We design each sequence to follow one of three behavioural patterns. In \textit{stable routines}, the underlying preference remains consistent across all turns (e.g., Preference A $\rightarrow$ A $\rightarrow$ A). In \textit{temporary deviations}, a short-term change occurs but reverts to the original behaviour (e.g., A $\rightarrow$ A $\rightarrow$ B $\rightarrow$ A). In \textit{preference shifts}, the user permanently transitions to a new preference (e.g., A $\rightarrow$ A $\rightarrow$ B $\rightarrow$ B). Examples of these sequences are provided in Appendix~\ref{app:longitudinal_examples}.

\paragraph{Evaluation Metrics}
Continual preference adaptation requires balancing the stability-plasticity trade-off, where a system must preserve previously learned preferences while remaining responsive to new user behaviour \cite{lai2025pareto}. Following this principle, we evaluate preference adaptation accordingly using four complementary metrics. \textit{Preference Consistency} is evaluated on stable sequences and measures whether the system reproduces the same preference under semantically similar commands. \textit{Recovery Rate} is evaluated on deviation sequences and measures whether the system correctly returns to the original preference following a temporary perturbation. This evaluates the robustness to noise. \textit{Adaptation Success} and \textit{Adaptation Delay} are evaluated on preference shift sequences. They measure whether and how quickly the system aligns with a new preference.

\subsubsection{Results}

\begin{figure}[h]
    \centering
    \includegraphics[width=0.7\linewidth]{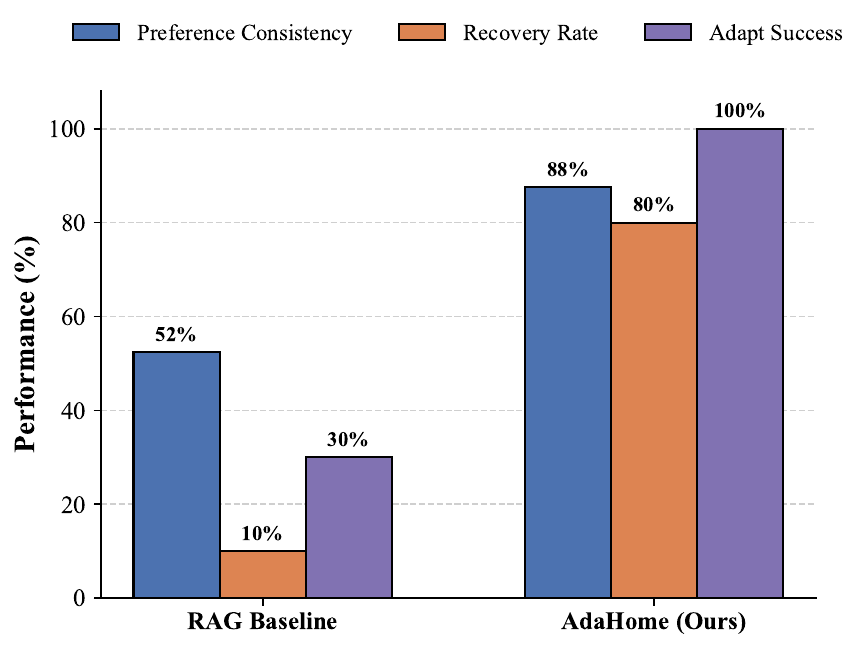}
    \caption{Multi-turn preference adaptation performance: AdaHome vs. RAG }
    \label{fig:pref_adaptation_result}
    \Description{xx}
\end{figure}

The results are presented in Figure~\ref{fig:pref_adaptation_result}. The RAG baseline achieves moderate stable accuracy (52.5\%), indicating that it can partially retain consistent preferences under repeated interactions. However, it exhibits a very low recovery rate (10.0\%), suggesting that it is highly sensitive to recent interactions and tends to overwrite previously learned preferences when exposed to temporary deviations. Although RAG adapts relatively quickly (2.0 turns on average to converge to a new preference), its low adaptation success (30.0\%) indicates that these updates are often inconsistent and do not reliably reflect true preference shifts.

In contrast, AdaHome achieves high stable accuracy (87.5\%), strong recovery (80.0\%), and a 100.0\% adaptation success. While adaptation is slightly slower (2.6 turns on average to converge), the updates are more reliable. This demonstrates that the system is capable of preserving previously learned preferences and effectively adapting to new ones over time. Compared to RAG, AdaHome provides a more balanced trade-off between stability and plasticity.

\subsection{Ablation Study}

We conduct an ablation study to examine the contribution of key components in AdaHome’s preference adaptation mechanism. We evaluate the role of the mechanism by removing memory entirely, and further isolate the effect of preference extraction within the mechanism. All experiments use the same dataset and evaluation metrics as described in the preference adaptation analysis. We compare three configurations:

\begin{itemize}
    \item \textbf{No Memory}: Actions are generated independently at each turn without access to prior interactions.
    
    \item \textbf{AdaHome (No Extractor)}: The preference mechanism applied directly on raw user commands without preference extractor.
    
    \item \textbf{AdaHome (Extractor)}: The full system with preference extractor.
\end{itemize}

\subsubsection{Results}

\begin{figure}[h]
    \centering
    \includegraphics[width=0.7\linewidth]{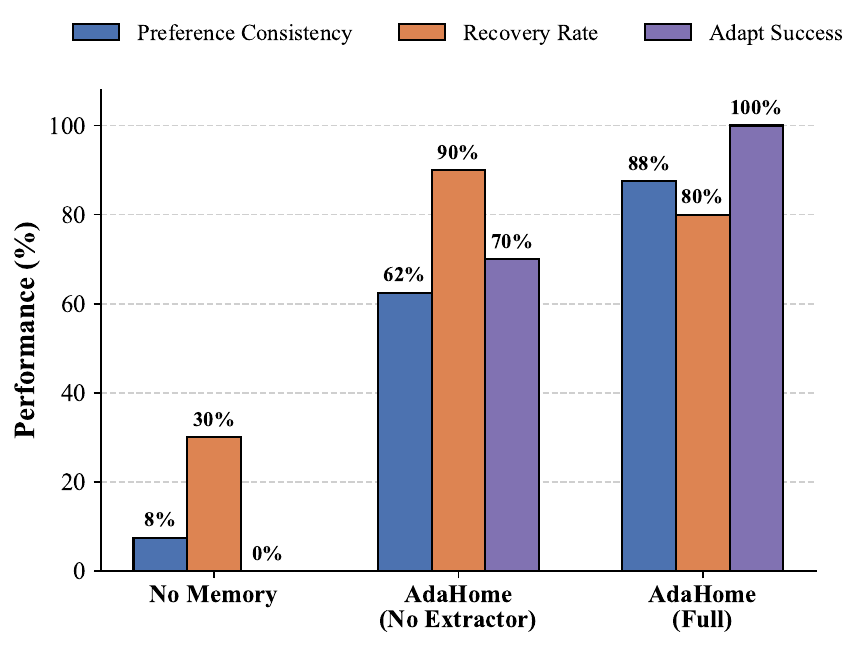}
    \caption{Ablation of AdaHome components: effect of memory and preference extraction.}
    \label{fig:ablation}
    \Description{xx}
\end{figure}

Figure~\ref{fig:ablation} presents the performance of all configurations. The No Memory baseline performs poorly across all metrics, achieving only 7.5\% stable accuracy and failing to adapt to preference shifts. This confirms that preference adaptation cannot be achieved without leveraging interaction history. 

AdaHome without preference extraction achieves the highest recovery rate (90.0\%) and improved adaptation success (70.0\%), reflecting strong responsiveness to recent user interactions. Operating directly on raw user commands makes the system highly sensitive to variations. For example, consider a temporary deviation sequence (movie night $\rightarrow$ let's watch a film $\rightarrow$ cinema mode $\rightarrow$ movie night). Without preference extraction, the system treats each command independently, allowing it to quickly switch behaviour in response to the most recent input, which leads to rapid recovery when the user returns to the original preference. However, this reactivity comes at the cost of slower adaptation (3.1 turns on average to converge). Variations in natural language expressions dilute semantic similarity, requiring more observations to reliably distinguish temporary deviations from true preference shifts.

Introducing preference extraction improves both adaptation reliability and efficiency. By extracting structured preference representations, the extractor aligns semantically equivalent inputs in the embedding space, enabling more consistent similarity matching. Compared to the variant without extraction, the full system achieves higher stable accuracy and adaptation success while reducing adaptation delay. This demonstrates that preference extraction reduces semantic noise and enables more accurate and consistent preference updates.

\section{Limitations and Future Work}

Our evaluation focuses on multi-turn preference adaptation over repeated interactions. While this demonstrates the system’s ability to adapt to evolving user behaviour, long-term user studies are needed to assess robustness under sustained usage. In addition, the current preference adaptation mechanism only models binary device states, and extending it to handle continuous device attributes is an important direction for future work. AdaHome also addresses cold-start scenarios by falling back to LLM reasoning. While this ensures functional behaviour from the outset, reducing reliance on early user feedback and accelerating personalization remain open challenges. Finally, our current experiments focus on text-based interaction. Incorporating additional multimodal signals, such as speech or gesture, would be an important next step toward more flexible and natural smart home interaction.


\section{Conclusions}

We presented AdaHome, an adaptive smart home assistant for locally deployed small language models. By selectively applying lightweight reasoning based on command type, AdaHome achieves a stronger efficiency–accuracy trade-off than existing approaches under small model constraints. Our evaluation further shows that AdaHome supports more reliable multi-turn preference adaptation than a RAG baseline, without requiring prompt augmentation or model retraining. These findings highlight the potential of small language models to enable efficient and personalized smart home interaction.

\section{Safe and Responsible Innovation}
AdaHome is designed for privacy-preserving, safe, and reliable smart home interaction. The system runs entirely locally using small language models, avoiding transmission of sensitive user data to external servers. Potential risks include incorrect device activation due to misinterpretation of user intent and over-adaptation to short-term behaviour that may not reflect true preferences. To mitigate these risks, AdaHome employs schema-constrained validation to enforce valid device states and a user confirmation step for ambiguous commands. In addition, continual user feedback is used to refine preference representations and improve robustness over time.

\bibliographystyle{ACM-Reference-Format}
\bibliography{sample-base}

\onecolumn
\appendix
\section{Intent Classification Prompt}
\label{app:intent_classification}

\vspace{1ex}
\begin{tcolorbox}[title=Intent Classification Prompt, colback=gray!5,colframe=black, width=\textwidth, arc=2mm, boxrule=0.5pt]
\begin{lstlisting}[language={},basicstyle=\ttfamily\small]
You are the Intent Classifier for a smart-home assistant.

Input:
- User command: {user_command}

Classify the user command into exactly one category:
- DIRECT: explicit device-control instruction with a clearly specified action
  and target device (e.g., "turn on the living room light")
- INDIRECT: implicit but inferable intent that can be resolved from context (e.g., "it is too hot")
- AMBIGUOUS: subjective, preference-dependent, or routine-like request that may
  have multiple valid interpretations depending on the user (e.g., "I want to relax")

Respond ONLY with valid JSON. 
The "category" key must strictly be one of the following: DIRECT, INDIRECT, or AMBIGUOUS.

Example:
{"category": "DIRECT"}
\end{lstlisting}
\end{tcolorbox}
\vspace{5ex}

\section{Chain-of-Draft Prompt}
\label{app:cod}

\vspace{1ex}
\begin{tcolorbox}[title=Reasoning Planner Prompt, colback=gray!5,colframe=black, width=\textwidth, arc=2mm, boxrule=0.5pt]
\begin{lstlisting}[language={},basicstyle=\ttfamily\small]
You are the Reasoning Planner for a smart-home assistant.

Input:
- Device states: {devices}
- User command: {user_command}

Infer the necessary device state changes based on the user command. 
Think step by step, but keep only a minimal draft with at most 8 words.

Respond EXACTLY in the following format:
Draft: <concise reasoning draft>
Action: <valid JSON object with state changes>

Use only the device names provided in the device states.
\end{lstlisting}
\end{tcolorbox}
\vspace{5ex}

\section{Preference Extraction Prompt}
\label{app:preference_extraction}

\vspace{1ex}
\begin{tcolorbox}[title=Preference Extraction Prompt, colback=gray!5,colframe=black, width=\textwidth, arc=2mm, boxrule=0.5pt]
\begin{lstlisting}[language={},basicstyle=\ttfamily\small]
You are the Preference Extractor for a smart-home assistant.

Input:
- User command: {user_command}

Extract a concise 1-2 word core intent that captures the underlying user goal.

Example:
- "I want to relax" -> {"core_intent": "relax"}

Respond ONLY with valid JSON using this format:
{"core_intent": "..."}

\end{lstlisting}
\end{tcolorbox}
\vspace{5ex}

\section{LLM-as-Judge Evaluation} 
\subsection{Prompt}
\label{app:llm_judge_prompt}

\vspace{1ex}
\begin{tcolorbox}[title=LLM-as-Judge Evaluation Prompt, colback=gray!5,colframe=black, width=\textwidth, arc=2mm, boxrule=0.5pt]
\begin{lstlisting}[language={},basicstyle=\ttfamily\small, breaklines=true, breakatwhitespace=true]
You are a smart-home system evaluator.

Input:
- Device states: {devices}
- User Command: {user_command}
- Agent Actions: {output}

Evaluate whether the agent's actions successfully satisfy the user's goal.

Evaluation Criteria:
- Success (1): The actions logically and safely satisfy the user's intent given the initial state. Multiple valid solutions are acceptable.
- Failure (0): The actions fail to satisfy the intent or activate unrelated devices.

Instructions:
- Provide a concise reasoning (1-2 sentences).
- Then output a final binary score.

Respond ONLY with valid JSON using this format :
{"reasoning": "...", "success": 1}
\end{lstlisting}
\end{tcolorbox}
\vspace{5ex}

\subsection{Example Outputs}
\label{app:llm_judge_example}
\vspace{1ex}
\begin{tcolorbox}[title=Success Example, colback=gray!5, colframe=black]

\textbf{Command:} time for bed \\
\textbf{Agent Action:} \\
\texttt{\{"living\_room\_light": \{"status": "off"\}, "bedroom\_light": \{"status": "off"\}\}} \\

\textbf{Judge Output:} \\
\texttt{\{"reasoning": "Turning off the lights creates a suitable environment for sleep and aligns with the user's intent.", "success": 1\}}

\end{tcolorbox}

\vspace{2ex}

\begin{tcolorbox}[title=Failure Example, colback=gray!5, colframe=black]

\textbf{Command:} movie night \\
\textbf{Agent Action:} \\
\texttt{\{"living\_room\_light": \{"status": "on"\}\}} \\

\textbf{Judge Output:} \\
\texttt{\{"reasoning": "Turning on the light alone does not capture the typical environment expected for a movie night, such as turning on the television.", "success": 0\}}

\end{tcolorbox}
\vspace{5ex}

\section{Smart Home Setting}
\label{app:smart_home_setting}

\begin{table*}[hbt!]
\centering
\caption{Smart home device schema}
\label{tab:device_schema}
\resizebox{0.7\textwidth}{!}{%
\begin{tabular}{ccl}
\hline
\textbf{Device} & \textbf{Type} & \textbf{Attributes} \\
\hline
living\_room\_light & Lighting & status, brightness \\
bedroom\_light & Lighting & status, brightness \\
living\_room\_heater & Climate & status, temperature \\
bedroom\_heater & Climate & status, temperature \\
living\_room\_ac & Climate & status \\
bedroom\_ac & Climate & status \\
living\_room\_television & Appliance & status \\
coffee\_maker & Appliance & status \\
air\_purifier & Appliance & status \\
bedroom\_blinds & Fixture & status (open/closed) \\
robot\_vacuum & Appliance & status (cleaning/docked) \\
front\_door\_lock & Security & status (locked/unlocked) \\
\hline
\end{tabular}%
}
\end{table*}
\vspace{5ex}

\section{Evaluation Dataset}
\label{app:dataset}
The evaluation dataset is grouped into three categories: direct, indirect, and ambiguous commands. Direct commands are evaluated using exact match against predefined target states, while indirect and ambiguous commands are assessed using an LLM-as-a-Judge due to the presence of multiple valid interpretations.

\subsection{Direct Commands}

\begin{table*}[hbt!]
\centering
\caption{Direct commands and their expected target states.}
\resizebox{0.9\textwidth}{!}{%
\begin{tabular}{ll}
\toprule
\textbf{User Command} & \textbf{Expected Target State} \\
\midrule
turn on the living room television & \texttt{\{"living\_room\_television": \{"status": "on"\}\}} \\
vacuum the bedroom & \texttt{\{"robot\_vacuum": \{"status": "cleaning"\}\}} \\
lock up the house & \texttt{\{"front\_door\_lock": \{"status": "locked"\}\}} \\
brew some coffee & \texttt{\{"coffee\_maker": \{"status": "on"\}\}} \\
turn off the bedroom light & \texttt{\{"bedroom\_light": \{"status": "off"\}\}} \\
turn on the air purifier & \texttt{\{"air\_purifier": \{"status": "on"\}\}} \\
close the bedroom blinds & \texttt{\{"bedroom\_blinds": \{"status": "closed"\}\}} \\
unlock the front door & \texttt{\{"front\_door\_lock": \{"status": "unlocked"\}\}} \\
turn on the living room ac & \texttt{\{"living\_room\_ac": \{"status": "on"\}\}} \\
switch off the living room television & \texttt{\{"living\_room\_television": \{"status": "off"\}\}} \\
turn the living room heater off & \texttt{\{"living\_room\_heater": \{"status": "off"\}\}} \\
open the bedroom blinds & \texttt{\{"bedroom\_blinds": \{"status": "open"\}\}} \\
turn on the bedroom ac & \texttt{\{"bedroom\_ac": \{"status": "on"\}\}} \\
dock the robot vacuum & \texttt{\{"robot\_vacuum": \{"status": "docked"\}\}} \\
turn the bedroom heater on & \texttt{\{"bedroom\_heater": \{"status": "on"\}\}} \\
stop the coffee maker & \texttt{\{"coffee\_maker": \{"status": "off"\}\}} \\
dim the living room light & \texttt{\{"living\_room\_light": \{"status": "on", "brightness": "low"\}\}} \\
lock the front door & \texttt{\{"front\_door\_lock": \{"status": "locked"\}\}} \\
turn the living room ac off & \texttt{\{"living\_room\_ac": \{"status": "off"\}\}} \\
set the bedroom light to high & \texttt{\{"bedroom\_light": \{"status": "on", "brightness": "high"\}\}} \\
start the robot vacuum & \texttt{\{"robot\_vacuum": \{"status": "cleaning"\}\}} \\
turn on the bedroom light & \texttt{\{"bedroom\_light": \{"status": "on"\}\}} \\
switch on the coffee maker & \texttt{\{"coffee\_maker": \{"status": "on"\}\}} \\
turn off the air purifier & \texttt{\{"air\_purifier": \{"status": "off"\}\}} \\
close the bedroom blinds now & \texttt{\{"bedroom\_blinds": \{"status": "closed"\}\}} \\
set the living room heater to high & \texttt{\{"living\_room\_heater": \{"status": "on", "temperature": "high"\}\}} \\
turn on the living room light & \texttt{\{"living\_room\_light": \{"status": "on"\}\}} \\
turn off the living room light & \texttt{\{"living\_room\_light": \{"status": "off"\}\}} \\
open the blinds in the bedroom & \texttt{\{"bedroom\_blinds": \{"status": "open"\}\}} \\
shut down the bedroom ac & \texttt{\{"bedroom\_ac": \{"status": "off"\}\}} \\
\bottomrule
\end{tabular}%
}
\end{table*}

\newpage
\subsection{Indirect Commands}

\begin{table*}[hbt!]
\centering
\caption{Indirect commands.}
\resizebox{0.8\textwidth}{!}{%
\begin{tabular}{ll}
\toprule
\textbf{User Command} & \textbf{User Command} \\
\midrule
make it less chilly in here & let some natural light into the bedroom \\
it is freezing and dark & I forgot to secure the house \\
it is too hot & it's too dark to read in the living room \\
it is too bright in the bedroom & I want to watch a movie \\
I can't see anything in the living room & the air in the bedroom is stale \\
the air feels stuffy & I need some coffee to wake up \\
it is roasting in the bedroom & my feet are cold in the living room \\
the sun is glaring in my eyes in the bedroom & the morning sun is waking me up \\
it's too loud, turn that off & the floor is covered in crumbs \\
clean up this mess on the floor & make sure the house is safe for the night \\
I need a caffeine boost & it's boiling in the living area \\
make sure nobody can get in & I'm going to sleep, kill the bedroom lights \\
it's getting a bit warm in the living room & I can't breathe well with all this dust \\
can you clear the air in here & I spilled chips everywhere \\
I'm shivering & I'm freezing in the bedroom \\
\bottomrule
\end{tabular}%
}
\end{table*}

\subsection{Ambiguous Commands}

\begin{table*}[hbt!]
\centering
\caption{Ambiguous commands.}
\resizebox{0.5\textwidth}{!}{%
\begin{tabular}{ll}
\toprule
\textbf{User Command} & \textbf{User Command} \\
\midrule
make it cozy in here & I want to relax \\
set up for a party & workout session \\
movie night & i feel like reading a book \\
good morning & getting ready for work \\
time for bed & it's date night \\
I'm leaving for work & time to wake up \\
welcome home & i'm feeling sick today \\
time to study & let's have a lazy sunday \\
romantic dinner & focus mode \\
i have a migraine & the kids are asleep \\
let's do some yoga & i'm throwing a dinner party \\
gaming time & energy saving mode \\
house cleaning day & i need to concentrate \\
guests are arriving & post-workout cooldown \\
afternoon nap & family board game night \\
\bottomrule
\end{tabular}%
}
\end{table*}

\newpage
\section{Longitudinal Preference Dataset}
\label{app:pref_dataset}
This dataset is derived from the ambiguous command category in the evaluation dataset and is used to simulate longitudinal user interactions. It is constructed using a set of predefined ambiguous scenarios, each associated with multiple paraphrases and two alternative preference configurations (Preference A and Preference B). For each scenario, we generate interaction sequences by sampling paraphrases and assigning preference trajectories according to three behavioural patterns: stable routines, temporary deviations, and preference shifts. Table~\ref{tab:longitudinal_scenarios} summarizes the ambiguous scenarios, paraphrases, and corresponding preference configurations used to construct the dataset.

\subsection{Preference Scenarios}
\label{app:pref_scenarios}
\begin{table*}[hbt!]
\centering
\caption{Ambiguous scenarios, associated paraphrases, and preference configurations used to generate longitudinal interaction sequences. LR denotes living room and BR denotes bedroom.}
\label{tab:longitudinal_scenarios}
\resizebox{0.9\textwidth}{!}{%
\begin{tabular}{lp{6.5cm}ll}
\toprule
\textbf{Intent} & \textbf{Paraphrases} & \textbf{Preference A} & \textbf{Preference B} \\
\midrule
Cozy 
& \textit{make the living room cozy} \newline 
  \textit{make my room comfortable} \newline 
  \textit{make it comfy} \newline 
  \textit{cozy evening mode} 
& LR heater on 
& LR heater on + LR light off \\
\midrule

Leaving house 
& \textit{i am leaving the house} \newline 
  \textit{heading out now} \newline 
  \textit{goodbye house} \newline 
  \textit{i'm going out} 
& Lock front door + LR light off + BR light off 
& Lock front door + vacuum on + purifier on \\
\midrule

Movie night 
& \textit{movie night} \newline 
  \textit{let's watch a film} \newline 
  \textit{cinema mode} \newline 
  \textit{time for a movie} 
& LR TV on + LR light on 
& LR TV on + LR light off \\
\midrule

Focus 
& \textit{i need to focus} \newline 
  \textit{time to study} \newline 
  \textit{help me concentrate} \newline 
  \textit{i need to get work done quietly} 
& LR light on + LR TV off 
& LR light on + purifier on \\
\midrule

Nap 
& \textit{taking a nap} \newline 
  \textit{i need to sleep for a bit} \newline 
  \textit{i want to rest} \newline 
  \textit{going to take a quick nap} 
& BR blinds closed + BR light off 
& BR light off \\
\bottomrule
\end{tabular}%
}
\end{table*}

\subsection{Example Interaction Sequences}
\label{app:longitudinal_examples}

\vspace{1ex}
\begin{tcolorbox}[title=Example: Stable Routine (Movie Night), colback=gray!5, colframe=black]

\textbf{Pattern:} A $\rightarrow$ A $\rightarrow$ A $\rightarrow$ A $\rightarrow$ A $\rightarrow$ A $\rightarrow$ A $\rightarrow$ A \\

\textbf{Intent:} Movie night \\

\textbf{Turns:}
\begin{itemize}
\item T1: movie night $\rightarrow$ TV on + light on
\item T2: let's watch a film $\rightarrow$ TV on + light on
\item T3: movie night $\rightarrow$ TV on + light on
\item T4: movie night $\rightarrow$ TV on + light on
\item T5: movie night $\rightarrow$ TV on + light on
\item T6: movie night $\rightarrow$ TV on + light on
\item T7: movie night $\rightarrow$ TV on + light on
\item T8: time for a movie $\rightarrow$ TV on + light on
\end{itemize}

\end{tcolorbox}

\vspace{1ex}
\begin{tcolorbox}[title=Example: Temporary Deviation (Cozy), colback=gray!5, colframe=black]

\textbf{Pattern:} A $\rightarrow$ A $\rightarrow$ A $\rightarrow$ A $\rightarrow$ A $\rightarrow$ B $\rightarrow$ A $\rightarrow$ A \\

\textbf{Intent:} Cozy \\

\textbf{Turns:}
\begin{itemize}
\item T1: cozy evening mode $\rightarrow$ heater on
\item T2: make my room comfortable $\rightarrow$ heater on
\item T3: make it comfy $\rightarrow$ heater on
\item T4: cozy evening mode $\rightarrow$ heater on
\item T5: make my room comfortable $\rightarrow$ heater on
\item T6: cozy evening mode (I have a headache) $\rightarrow$ heater on + light off
\item T7: make my room comfortable $\rightarrow$ heater on
\item T8: make my room comfortable $\rightarrow$ heater on
\end{itemize}

\end{tcolorbox}

\vspace{3ex}
\begin{tcolorbox}[title=Example: Preference Shift (Movie Night), colback=gray!5, colframe=black]

\textbf{Pattern:} A $\rightarrow$ A $\rightarrow$ A $\rightarrow$ A $\rightarrow$ B $\rightarrow$ B $\rightarrow$ B $\rightarrow$ B \\

\textbf{Intent:} Movie night \\

\textbf{Turns:}
\begin{itemize}
\item T1: time for a movie $\rightarrow$ TV on + light on
\item T2: time for a movie $\rightarrow$ TV on + light on
\item T3: cinema mode $\rightarrow$ TV on + light on
\item T4: time for a movie $\rightarrow$ TV on + light on
\item T5: cinema mode $\rightarrow$ TV on + light off
\item T6: time for a movie $\rightarrow$ TV on + light off
\item T7: cinema mode $\rightarrow$ TV on + light off
\item T8: let's watch a film $\rightarrow$ TV on + light off
\end{itemize}

\end{tcolorbox}

\end{document}